\pdfoutput=1
\documentclass[11pt]{article}

\usepackage[final]{acl}

\usepackage{times}
\usepackage{latexsym}
\usepackage{booktabs}
\usepackage{subcaption}
\usepackage[normalem]{ulem}
\usepackage{amsmath}
\useunder{\uline}{\ul}{}
\usepackage{listings}

\usepackage{xcolor}
\definecolor{codebg}{RGB}{255, 249, 230}
\lstdefinestyle{pythonstyle}{
    language=Python,
    basicstyle=\ttfamily\footnotesize,  
    backgroundcolor=\color{codebg}, 
    breaklines=true,
    commentstyle=\color{green!60!black},
    keywordstyle=\color{blue},
    stringstyle=\color{red},
    numbers=left,
    numberstyle=\tiny\color{gray},
    numbersep=10pt,
    tabsize=4,
    showspaces=false,
    showstringspaces=false,
    frame=single,                      
    framesep=3pt,                     
    xleftmargin=10pt,                 
    xrightmargin=0pt                  
}

\usepackage[T1]{fontenc}
\makeatletter
\renewcommand*{\@fnsymbol}[1]{\ensuremath{\ifcase#1\or \dagger\or \ddagger\or
   \mathsection\or \mathparagraph\or \|\or **\or \dagger\dagger
   \or \ddagger\ddagger \else\@ctrerr\fi}}
\makeatother
\usepackage[utf8]{inputenc}

\usepackage{microtype}

\usepackage{inconsolata}

\usepackage{graphicx}

\newcommand*\samethanks[1][\value{footnote}]{\footnotemark[#1]}

\title{Standardizing the Measurement of Text Diversity: \\ A Tool and Comparative Analysis}

\author{\textbf{Chantal Shaib$^1$\thanks{Partial work completed while at Adobe Research.}}\quad\quad
\textbf{Venkata S Govindarajan}$^4$\quad\quad
\textbf{Joe Barrow$^3$\samethanks}\quad\quad
\textbf{Jiuding Sun$^1$}\quad\quad \\
\textbf{Alexa F. Siu$^2$}\quad\quad
\textbf{Byron C. Wallace$^1$}\quad\quad
\textbf{Ani Nenkova$^2$}\\
$^1$Northeastern University, $^2$Adobe Research, $^3$Pattern Data, $^4$Ithaca College\\
\small\texttt{shaib.c@northeastern.edu}
}

\begin{document}
\maketitle

\begin{abstract}
The diversity across outputs generated by LLMs shapes perception of their quality and utility. High lexical diversity is often desirable, but there is no standard method to measure this property.
Templated answer structures and ``canned'' responses across different documents are readily noticeable, but difficult to visualize across large corpora. 
This work aims to standardize measurement of text diversity. 
Specifically, we empirically investigate the convergent validity of existing scores across English texts, and release \texttt{diversity}, an open-source Python package\footnote{\hyperlink{https://pypi.org/project/diversity/}{https://pypi.org/project/diversity/} \label{pkg}} for measuring and extracting repetition in text. We also build a platform\footnote{\url{https://ai-templates.app}\label{site}} based on \texttt{diversity} for users to interactively explore repetition in text.  
We find that fast compression algorithms capture information similar to what is measured by slow-to-compute $n$-gram overlap homogeneity scores. 
Further, a combination of measures---compression ratios, self-repetition of long $n$-grams, and Self-BLEU
---are sufficient to report, as they have low mutual correlation with each other. 
\end{abstract}

\section{Introduction}

LLM-generated texts are typically evaluated with respect to accuracy or factuality, e.g., as measured via entailment \citep{tang-etal-2023-understanding}, or text quality aspects such as coherence and fluency (e.g., estimated using LLMs as evaluators  \citealt{Liu2023GEvalNE}). 
When reference summaries are available, the similarity of generated outputs to these is also often measured (e.g., via ROUGE; \citealp{lin-och-2004-automatic}). 
A complementary dimension of model performance is \emph{diversity}, or how much ``boilerplate'' content is repeated \emph{across} LLM outputs. 


There is a distinct lack of standardization in reporting diversity in ML datasets \citep{pmlr-v235-zhao24a}.  
We address this by introducing \texttt{diversity}, an open-source Python package for evaluating text diversity,\footref{pkg}
along with a web-based UI that allows users to visualize repetition in their corpus,\footref{site} 
providing an intuitive, efficient tool for text analysis that permits: (i) Viewing repetitive text and Part-of-Speech $n$-grams; (ii) Quickly computing diversity metrics, and; (iii) Interactively highlighting and matching repetition in documents. 
Both the package\footnote{ \url{https://github.com/cshaib/diversity}} and UI code\footnote{\url{https://github.com/cshaib/diversity_demo}} are open-sourced under the Apache 2.0 license.

We run existing diversity metrics over English language outputs from several LLMs to identify a few (mostly) independent scores that characterize repetition. 
We also examine diversity in downstream datasets such as instruction tuning. 
Finally, we show that \emph{compression ratio}---compressed over original texts size---is a fast, easy to compute score sufficient to capture the information in all token/type ratio related alternatives. 
But we emphasize text length as an important confounder when assessing diversity: No reliable conclusions can be drawn without taking this into consideration.  
 
\begin{figure*}
\centering
        \centering
        \includegraphics[width=.88\textwidth]{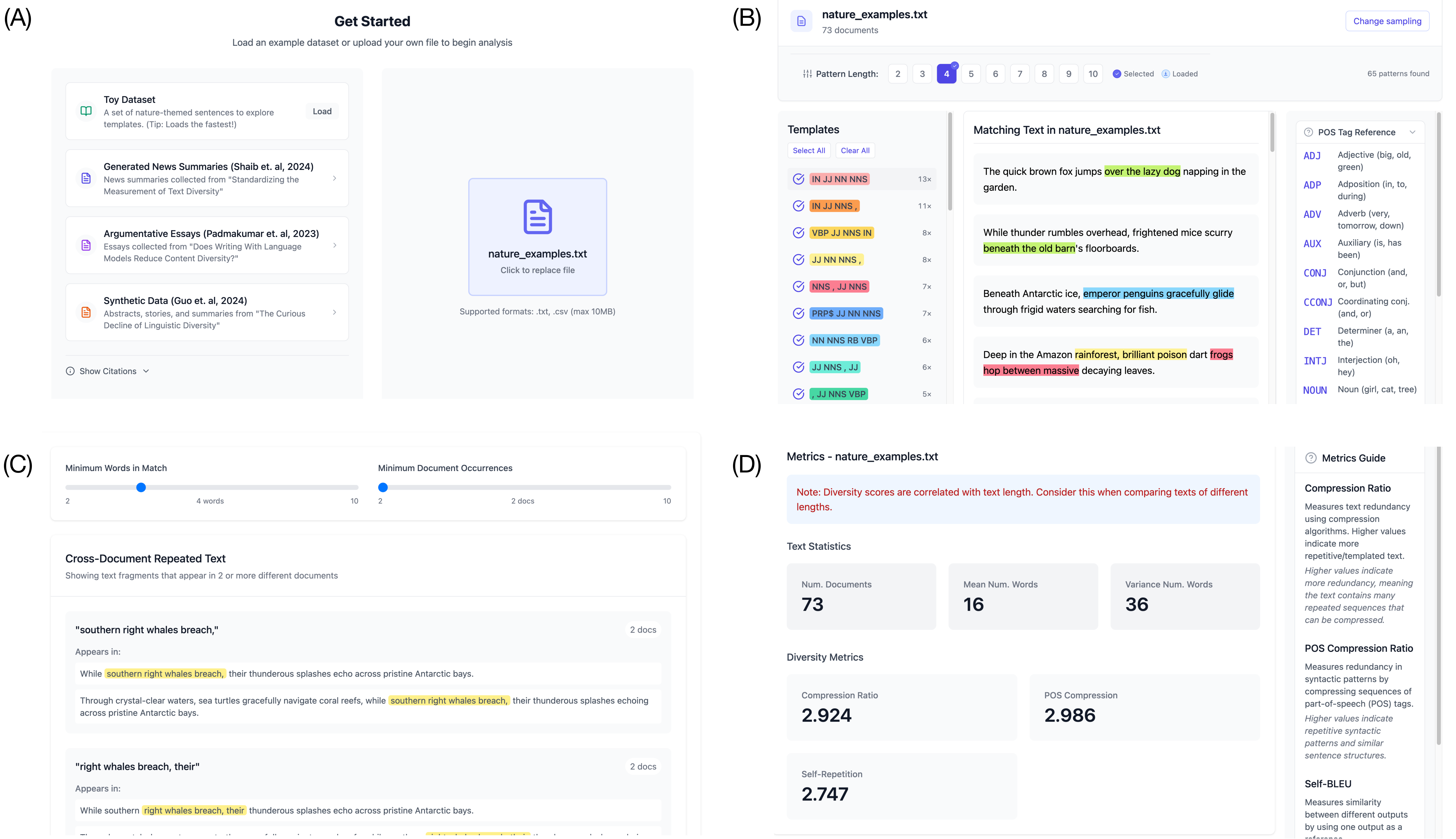}
        \caption{\textbf{(A)} Users start by uploading their own dataset on the right or by selecting one of the existing demo datasets on the left. Once uploaded, users can \textbf{(B)} interactively visualize part-of-speech patterns in the data, \textbf{(C)} interactively search for exact repeated text matches, and \textbf{(D)} calculate lexical diversity metrics.}
        \vspace{-.75em}
        
    \label{fig:ui_full}
\end{figure*}

Our contributions are as follows. (1) We introduce \texttt{diversity}, a Python package implementing diversity metrics. (2) We host and release source code to a user interface to explore repetition and diversity. (3) We evaluate the convergent validity of existing lexical diversity metrics and highlight compression ratios as efficient measures of diversity. 

\section{Related Work}
\vspace{-0.65em}
Lack of diversity in text may result from repetition of lengthy strings or owe to subtle distributional patterns 
\citep{holtzman2019curious,DBLP:journals/tacl/MeisterWC22,meister-etal-2023-locally}. 
We focus on scores that aim to capture overt repetition across outputs, and leave for future work similar analysis of semantic and structural diversity scores \citep{bar-etal-2012-text, shaib-etal-2024-detection}.
Conditional generation tasks such as image captioning have offered observations regarding the diversity of produced texts. Prior work has shown that models tend to repeat the same text for different contexts in these tasks \cite{li-etal-2016-diversity,devlin-etal-2015-language}. 
Self-repetition~\citep{salkar2022self}---exact repetition of the same $n$-gram ($n \geq 4$) across outputs---is a practical way of measuring repetition in lengthy outputs.
In such cases repetition is common, especially relative to training data \citep{wang-etal-2023-automated}. 

We discuss several metrics but it is unclear which of these to use when, and how to efficiently visualize lower diversity in text-only tasks. 
Further, prior work has shown that human judgments of diversity are difficult to reliably collect. Humans tend to implicitly conflate quality of text with its diversity, and it can be difficult to separate content and lexical diversity in such assessments \citep{tevet-berant-2021-evaluating}. 
We design an interactive tool to allow users to browse highlighted instances of ``lower diversity'' text (Figure~\ref{fig:ui_full} (B)). 
\subsection{A Smorgasbord of Text Diversity Scores}
\label{sec:ld}
Scores used to measure diversity across a corpus of texts derive from two core ideas: Computing average similarity between pairs of outputs produced by the same model for different inputs, and computing variants of token/type ratio. 
The former are adapted from common approaches to reference-based text generation using standard measures of pairwise similarity; the latter track the diversity of vocabulary measured as the ratio of unique words to total words produced, with outputs from a model concatenated into a single text. 
We first describe each score, and then  
present insights regarding their mutual redundancy. 
All scores are defined for a set of generated texts $D$, each conditioned on its respective input.

\vspace{0.38em}
\noindent{\bf{Self-BLEU}} The quality of text in machine translation, summarization, and image captioning is often reported in terms of overlap with a reference text.
This idea can be adapted to measure diversity across different outputs by using one generated text as a ``reference'' and measuring the similarity of other outputs against this.
Self-BLEU measures similarity between all text pairs in $D$ using BLEU \citep{Zhu2018TexygenAB}. 
BLEU could be replaced with other similarity scores, e.g., ROUGE-L or BERTScore. 
These variants are called \textbf{homogenization scores} and have recently been used to compare the diversity of texts produced under several conditions \citep{Padmakumar2023DoesWW}.

\vspace{0.38em}
\noindent{\bf{Homogenization Score (ROUGE-L)}} All homogenization scores calculate an aggregate similarity across pairs of examples (Equation \ref{eq:homogeneity score}). 
Here the similarity score of choice is ROUGE-L \citep{lin-och-2004-automatic},  which quantifies overlap in terms of longest common sub-sequences between all pairs of text in a  
corpus instead of the fixed $n$-gram size used in other ROUGE variants:

\begin{equation}
    \text{hom}(D) = \frac{1}{|D|-1} \sum_{d, d' \in D; \: d \neq d'} \text{sim}(d, d')
    \label{eq:homogeneity score}
\end{equation}

\vspace{0.38em}
\noindent{\bf{Homogenization Score (BERTScore)}} 
This homogenization score uses BERTScore to measure similarity between documents in Equation \ref{eq:homogeneity score}. 
Unlike the other scores, it does not count the repetition of specific tokens, but instead uses BERT embeddings to (ideally) capture ``semantic'' similarity beyond verbatim $n$-gram matches. 

\vspace{0.38em}
\noindent{\bf{Self-repetition Score}} Self-repetition measures the tendency of LMs to repeat long $n$-grams across different outputs \citep{salkar2022self}.

\begin{equation}
\text{SRS}(d) = \log \left ( \sum\limits_{i=1}^k N_i + 1 \right)
    \label{eq:self-rep}
\end{equation}

\noindent Where $k$ is the total number of 4-grams in a single document $d$ and $N_i$ the number of other summaries in which 4-gram $i$ appears. The final score is the sum of $\text{SRS(}d\text{)}$ divided by $|D|$. 

\vspace{0.38em}
\noindent{\bf{Moving Average Token-Type Ratio}}
The token-type ratio for a text is the 
unique token count divided by the total token count. 
This metric captures the repetition of a given word in segments of text and does not explicitly account for longer repeated sequences \citep{Covington2010CuttingTG}.

\vspace{0.38em}
\noindent{\bf{$N$-Gram Diversity Score}} NGD extends the idea of token-type ratio to longer $n$-grams \citep{Padmakumar2023DoesWW, meister-etal-2023-locally, li-etal-2023-contrastive}, taking a ratio of unique to all $n$-gram counts:

\begin{equation}
    \text{NGD}(D) = \sum\limits_{n=1}^4\frac{\text{\# unique }n\text{-grams in } D\oplus}{\text{\# }n\text{-grams in } D\oplus}
    \label{eq:ngram diversity score}
\end{equation}

\noindent Where $D\oplus$ denotes the dataset $D$  concatenated into a single string. 
We use four as the maximum $n$-gram length. 
This method captures repeated \emph{sequences} in addition to single token diversity.


\subsection{Compression Ratios for Diversity}
\vspace{0.38em}
\noindent{\bf{Compression Ratios (CRs)}}
The diversity scores introduced so far are all a function of the number of repeated substrings across outputs. We use gZip to compress the concatenated text of all outputs generated by a model. 
CR is then the ratio between the size of the compressed file to that of the original. 
High CRs imply more redundancy:

\begin{equation}
    \label{eq:compression}
    \text{CR}(D) = \frac{\text{size of } D\oplus}{\text{compressed size of } D\oplus}
\end{equation}

\vspace{0.38em}
\noindent{\bf{Part-of-Speech Compression Ratio}} To capture repeated syntactic patterns, we also compute compression ratios for part-of-speech (POS) tag sequences. We use the NLTK POS tagger \footnote{\url{https://www.nltk.org/api/nltk.tag.html}} and the Penn Treebank set of 36 tags.
\section{Evaluating Repetition with \texttt{diversity}} 

\subsection{Design of the Diversity Package}
The \texttt{diversity} package incorporates measures of diversity including lexical, syntactic, and semantic diversity. 
For lexical/syntactic diversity, we use NLTK \citep{bird-loper-2004-nltk} and SpaCY \cite{Honnibal_spaCy_Industrial-strength_Natural_2020} to tag text with parts of speech and extract $n$-grams.
We also include implementations of \textbf{embedding-based} measures \citep{Cox_2021}, and \textbf{QUDSim} \citep{namuduri2025qudsim}. 
Users can install the package via pip (assuming Python 3.10+) and can calculate various diversity metrics over a list of texts as follows: 

\begin{lstlisting}
from diversity import *
    
text = ["I enjoy walking with my cute dog...", "I enjoy walking outside with...", "I enjoy jogging on a sunny..."]
# compression ratios
cr = compression_ratio(text, 'gzip')
cr_pos = compression_ratio(get_pos(text)[1], 'gzip')
# homogenization scores
hs_rougel = homogenization_score(text, 'rougel')
hs_bert = homogenization_score(text,  'bertscore')
self_bleu = homogenization_score(text, 'bleu')
# other
self_rep = self_repetition_score(text)
nds = ngram_diversity_score(text, n=4)

# Embedding-based
rc = remote_clique(text, model="Qwen/Qwen3-Embedding-0.6B", verbose=False)

cd = chamfer_dist(text, model="Qwen/Qwen3-Embedding-0.6B", verbose=False)

# QUDSim
key = os.environ.get("OPENAI_API_KEY")  # requires an OpenAI key
qud_alignment = qudsim(text, key=key)   # list of QUD-based alignments/scores
\end{lstlisting}

Users can extract PoS patterns using the \texttt{\textcolor{darkblue}{extract\_patterns}} function by specifying the $n$-gram length to search for, and the top\_n most repeated $n$-grams to return: 
\begin{lstlisting}
extract_patterns(text, n=5, top_n=100)
\end{lstlisting}
This returns a Python dictionary where the keys are the part-of-speech n-grams and the values are the raw text n-grams matching those patterns. The pattern matches are based on the frequency seen across the entire dataset, i.e., a part-of-speech pattern is only a pattern if it appears in more than 2 texts in the original input. Default values consider the top 100 part-of-speech patterns (sorted by frequency).

Then, using \texttt{\textcolor{darkblue}{match\_patterns}}, a user can identify all patterns in a single text from the input: 

\begin{lstlisting}
idx = 2
match_patterns(text[idx], patterns)
\end{lstlisting}

\noindent which returns a list of tuples containing the pattern and matched substring, respectively.
Many diversity metrics require pairwise comparisons. With larger datasets, this can become infeasible to compute (see Appendix~\ref{runtime}). We implement a few methods to increase efficiency: memoization of already computed pairs, and batch pattern searching in the UI. 

We also include a function for users to run all metrics and display them in a table to easily compare values: 

\begin{lstlisting}
compute_all_metrics(corpus=text)
\end{lstlisting}

\subsection{Metric Visualization and the Web UI}
The \texttt{diversity} Web UI offers the same functionality as the package via a no-code UI. 
Figure~\ref{fig:ui_full} shows the main pages of the site: users can begin by (A) either uploading their own text file for analysis or selecting one of the demo datasets provided on the site. Then, the user is prompted to select one of three types of analyses: either (B) to explore part-of-speech patterns, (C) to explore verbatim repeated text, or (D) to measure various diversity metrics of the dataset. 
Datasets are processed upon upload, and nothing is stored on the backend server aside from the existing demo datasets. 

\vspace{0.25em}
\noindent{\bf{(B) Templates}}
The templates tab allows users to explore extracted part-of-speech $n$-grams in their selected dataset. 
The left-most column displays pattern length of $n = [2, 10]$. The user can then scroll through all of the templates, select some or all, and see the highlighted text in the middle panel corresponding to the template. The templates are assigned a colour when selected to indicate the corresponding matched text. The right-most column provides a reference for all the part-of-speech tags from SpaCY. \footnote{\url{https://spacy.io/usage/linguistic-features}} 
The default pattern length is set to $n=4$. Other lengths will load when selected.

\vspace{0.25em}
\noindent{\bf{(C) Exact Match}} The exact matches tab allows a user to explore exact text matches in their dataset. The top provides two sliders: the left slider allows the user to set a string length to search for ($n = [2, 10]$), and the right the minimum number of documents in which the string must appear ($n = [2, 10]$). The minimum document occurrence slider defaults to 2. Once selected, the user can scroll through to see the repeated text in bold, and the full document text in which the string appears, as well as the number of documents. 

\vspace{0.25em}
\noindent{\bf{(D) Diversity Metrics}}
The diversity metrics tabs reports the recommended metrics from our evaluation: Compression Ratio, POS Compression, Self-BLEU, Self-Repetition, and Homogenization with BERTScore. We display these values alongside a guide to the metrics on the right-hand side. 

\subsection{Use-Cases}
Our implementation of compression ratios over PoS tags and tokens (along with BERTScore, Self-BLEU, and self-repetition) have already been used in prior works (by other groups) to evaluate diversity in model evaluation and alignment \citep{Lake2024FromDT, Moon2024DiffSLTED, fernandez2024divert}, and for reporting diversity over synthetic datasets \citep{Chang2024ScalingPL, Hastings2024UtilizingLL}. Due to its computational efficiency, compression ratios have also been used as optimization parameters in decoding strategies \citep{Lanchantin2025DiversePO}.

\citet{shaib-etal-2024-detection} use the pattern analysis in \texttt{diversity} to measure and evaluate the prevalence of syntactic patterns in LLMs. 
\citet{Wadhwa2025WhoTY} extract PoS patterns in distillation tasks for model attribution.
Further, insights from our evaluation of diversity metrics have informed how to report diversity with respect to text length and data sizes \citep{DBLP:journals/corr/abs-2311-09807, Hastings2024UtilizingLL}.

\section{Platform Evaluation: Comparative Analysis of Diversity Metrics}
\subsection{Data and Models}

We compute diversity scores for the outputs of six instruction tuned models on the CNN/DailyMail \citep{Hermann2015TeachingMT} and XSUM \citep{Narayan2018DontGM} English news summarization datasets: Llama-2 \citep{Touvron2023LLaMAOA}, GPT-4 \citep{OpenAI2023GPT4TR}, FlanT5-XXL \citep{Longpre2023TheFC}, StableLM \citep{alpaca, vicuna2023,gpt4all}, Mistral \citep{Jiang2023Mistral7}, and StableBeluga \citep{touvron2023llama, mukherjee2023orca}.\footnote{All models---except GPT-4---downloaded from {\sc HuggingFace} (\url{https://huggingface.co/models}).} 
We selected these models to cover a range of availability (open and closed), and architectures (encoder-decoder, decoder-only).
The lengths of texts vary considerably by source, for reference and model-produced text alike, so we also note average lengths when reporting diversity.

\begin{figure}
\centering
    \includegraphics[width=0.35\textwidth]{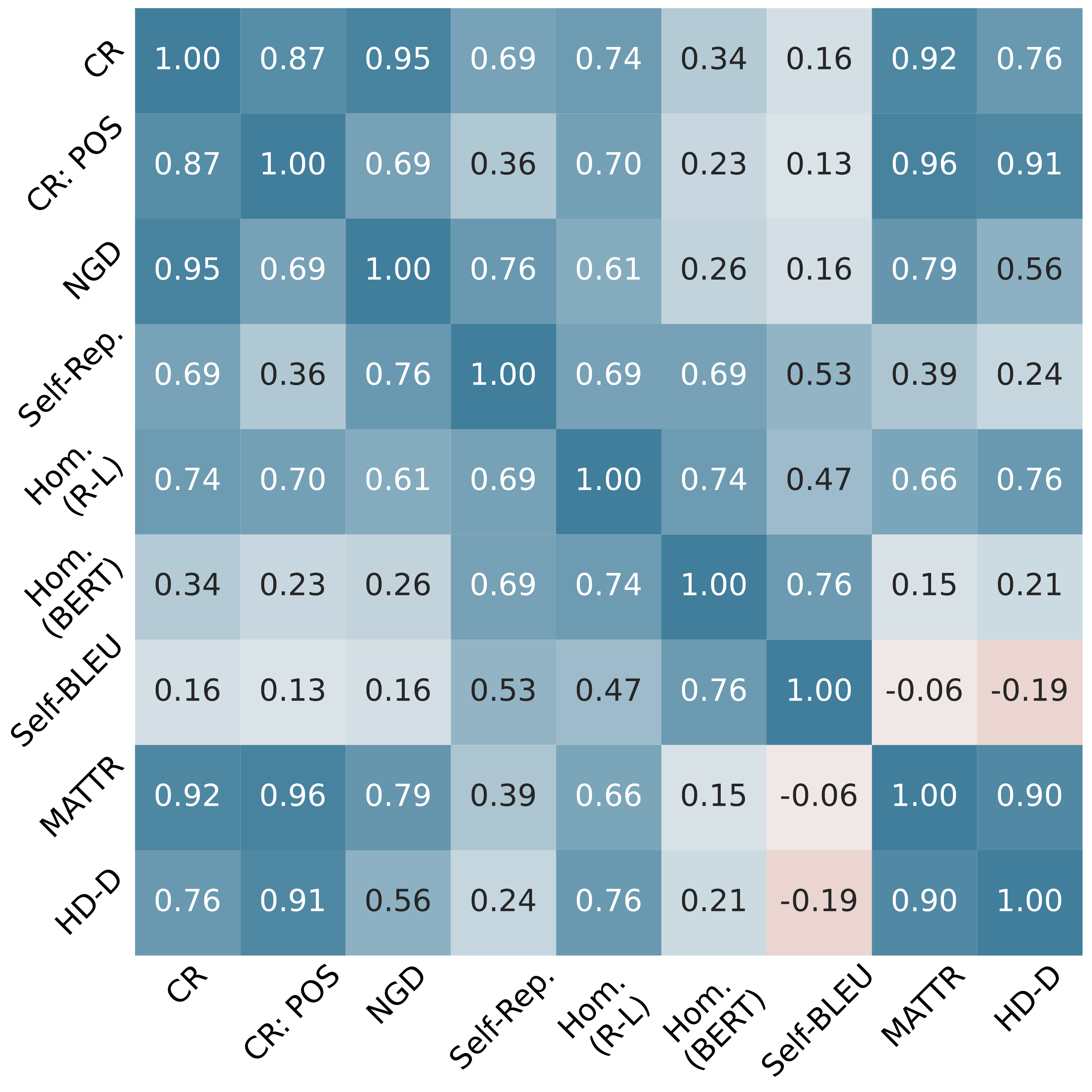}
    \caption{Correlations between diversity scores on CNN/DM. CR correlates strongly with most other metrics.}
    \vspace{-.75em}
    \label{fig:corr_cnn}
\end{figure}




\section{Text Length as a Confounder}

To keep compute time and costs  
manageable, we randomly sample 500 inputs from  
CNN/DailyMail and XSUM 
for analysis. 
Table~\ref{table:cnn_diversity} reports  
diversity scores for outputs generated by the six LLMs 
for these inputs.  
Table~\ref{table:cnn_diversity} (top) reports scores for human-written texts: The article given as input for summarization, the baseline summary comprising the first three sentences of the news article, and the reference summary. 
These scores serve as a reference point for the diversity scores of the models.

\begin{table*}
\begin{centering}
\resizebox{0.9\textwidth}{!}{
\begin{tabular}{@{}lllllllllll}
\toprule
\textbf{Model}                                                     &   \textbf{\begin{tabular}[c]{@{}l@{}}Avg.\\Length\end{tabular}}&\textbf{\begin{tabular}[c]{@{}l@{}}CR\\(↓) \end{tabular}}& \textbf{\begin{tabular}[c]{@{}l@{}}CR: POS\\(↓) \end{tabular}}& \textbf{\begin{tabular}[c]{@{}l@{}}NGD\\(↑) \end{tabular}} &  \textbf{\begin{tabular}[c]{@{}l@{}}Self-\\Rep. (↓) \end{tabular}}&\textbf{\begin{tabular}[c]{@{}l@{}}Hom. \\(R-L) (↓)\end{tabular}}& \textbf{\begin{tabular}[c]{@{}l@{}}Hom. \\(BERT) (↓)\end{tabular}}& \textbf{\begin{tabular}[c]{@{}l@{}}Self-\\BLEU (↓)\end{tabular}} & \textbf{\begin{tabular}[c]{@{}l@{}}MATTR\\(↑) \end{tabular}} & \textbf{\begin{tabular}[c]{@{}l@{}}HD-D\\(↑) \end{tabular}} \\\midrule
 Article                                                     &  452.25&2.615& 5.544& 2.637          & 6.216& 0.118                                                                  & 0.696                                                               & 0.003          & 0.837          &0.896           \\
 Article (Lead 3) &  75.87                                                                                      &2.369& 5.497& 3.041          & 4.276& 0.105                                                                  & 0.686                                                               & 0              & 0.856          &0.892           \\
 Reference                                                   &  51.78                                                                                      &2.277& 5.330& 3.164          & 3.842& 0.074                                                                  & 0.683                                                               & 0              & 0.875          &0.919           \\ \midrule
StableLM                                                    &   132.71&\textbf{2.724}& \textbf{5.940}&2.673 &  4.940&\textbf{0.126}                                                         & 0.689                                                               & 0.002          & 0.792 & 0.867  \\
 Mistral                                                     & 114.88& 2.499& \textbf{5.621}& 2.926          & 4.688& \textbf{0.123}                                                                  & \underline{\emph{0.697}}                                                               & \underline{\emph{0.036}}          & 0.831          &0.880            \\
 Llama-2                                                     & 106.52                                                                                     & \underline{\emph{2.543}}& \underline{\textbf{5.684}}& \underline{2.874}          & 4.159*& \underline{\textbf{0.125}}                                                                  & \underline{\emph{0.694}}                                                               & 0.001          & \underline{0.820}           &\underline{\emph{0.873}}           \\
 StableBeluga                                                & 91.17                                                                                      & 2.452& \underline{\textbf{5.644}}& 3.028          & \underline{4.467}& \textbf{0.121}                                                                  & \underline{\emph{0.702}}                                                      & \underline{\emph{0.047}} & 0.846          &0.889           \\
 FlanT5                                                      & 63.84& \underline{\textbf{2.453}}& \textbf{5.608}& \underline{\emph{2.939}}          & 3.608*& 0.084& 0.667& 0& \underline{\emph{0.833}}          &\underline{\textbf{0.887}}           \\
GPT-4                                                       &   55.4                                                                                       &2.361& \textbf{5.463}&\textbf{3.124}&  \underline{\emph{3.909}}&\underline{\emph{0.098}}                                                                  & \underline{\emph{0.684}}                                                               & \textbf{\underline{\emph{0.001}}}          & 0.853& \textbf{0.891}\\\midrule
\end{tabular}
}
\caption{Diversity scores for the CNN/Daily Mail dataset. Arrows indicate direction of \textit{more diversity}. Values indicating less diversity compared to at least one text source that produces longer human texts are bolded; models with scores that are less diverse than those from a model that produces longer summaries are underlined. An asterisk indicates a model more diverse than a shorter human text.} 
\vspace{-.75em}
\label{table:cnn_diversity}
\end{centering}

\end{table*}

One would expect that human-authored texts would be more diverse than those produced by LLMs (with the caveat that the texts were scraped from the web, and so may contain HTML and page layout artifacts which might be repetitive \citep{salkar2022self}). 
The human texts differ by length and the sources of longer texts appear to be less diverse. 

Text length as a confounder for diversity has been reported in prior work \citep{salkar2022self}, along with methods to account for this, e.g., sampling blocks of fixed size \citep{Covington2010CuttingTG}.  

All scores of the token/type ratio family are highly correlated with length, while the pairwise similarity ones are only moderately correlated. Self-BLEU has low correlation with length.

\section{Diversity of Model Summaries}

The confound of length complicates reporting. On 
CNN/DM 
(cf. Table \ref{table:cnn_diversity}) 
StableLM produces the longest summaries. All scores indicate that these are the least diverse, probably due to length. 
Three types of differences are marked in the tables. 
Model summaries that are shorter but less diverse than human summaries are marked in bold. 
Human texts here are written by journalists, so the expectation is that they would be more diverse. 
More bold entries in a column indicate that the score captures differences between human and machine diversity, a desirable trait. 
Underlined entries denote models that are less diverse than other models that produce longer summaries. 
The more underlined entries there are for a model, the more indicators there are that its output is less diverse.
Asterisks mark models that appear more diverse than a human text of shorter length.

The most interesting diversity scores are those that capture differences between human and automatically produced text. 
On the CNN/DM dataset, Hom. (BERT) and MATTR are the two scores that detect no differences between human and model texts.
Compression ratio for part of speech sequences is the score that identifies the most differences between human and model-generated text. 
Self-repetition stands out as the only score that identifies model generated text as more diverse on the CNN/DM dataset. From this analysis, CR:POS and self-repetition emerge as prime candidates of reportable scores, while Hom. BERT is less useful.

\section{Correlation Analysis}
We present three sets of correlation analyses between \emph{(i)} different diversity scores, \emph{(ii)} the same diversity score across datasets, and \emph{(iii)} diversity scores and standard reference-based evaluations. 
Despite the large number of diversity scores in our list, they all revolve around $n$-gram repetition. 
Do these capture different (complementary) information? 
To assess this we compute the correlations between all pairs of scores, reported in Figure~\ref{fig:corr_cnn}.

Compression ratio is highly to moderately correlated with other $n$-gram scores. 
The only weak correlations are with Self-BLEU and Hom. (BERT).
Given the degenerate behavior of Hom. (BERT) on the analysis of summaries, reporting Self-BLEU only is advisable. 
Finally, self-repetition is only moderately correlated with other scores, and is therefore informative to report.
Correlations are similar on XSUM summaries (Appendix~\ref{fig:corr_xsum}). 
\section{Truncating to Control for Length}
We truncate all summaries to the length of the shortest one produced by any source as a crude means to control scores for length.
The resulting scores are directly comparable (see Table \ref{fig:corr_xsum} in the Appendix).
CRs and Self-BLEU scores indicate that model-generated text is less diverse than human text. 
Hom. (BERT) scores barely vary across sources.
On the CNN/DM dataset, Self-BLEU indicates that Llama-2 and StableLM are the most repetitive models. 
CR also ranks these two models as the least diverse. The results are consistent on XSUM, but for that dataset Flan-T5 is also highly ranked and the most repetitive. 

Truncation to control for length is impractical for published research or leaderboards. Introducing a new source of texts would require recomputing scores for other sources for comparison, which is sometimes impossible (when outputs from other sources are not available). Future research might search for more practical alternatives. 

\section{Discussion and Recommendations}
The \texttt{diversity} package and platform provides a useful way to analyze and visualize diversity in datasets. Our analyses of metrics reveal that compression ratio (CR) is an excellent score to report, easy to compute and strongly correlated with other scores used in past work. CR of PoS sequences captures differences between human and model-generated text. Self-repetition focusses on repetition of longer $n$-grams across generations, and is only moderately correlated with CRs. Finally Self-BLEU is only weakly correlated with the previous three, so is a good complement score to report. We found BERTScore limited: It does not show differences between human and model-generated text and barely varies when adjusted for length. 

Length of the analyzed text has to be reported alongside all these scores. When length differs, scores are not meaningfully comparable. Truncating text is one way to control for this. Different random draws of the sample chosen to represent a dataset may differ in diversity, in turn leading to unwarranted conclusions. Truncating texts prevents quantifying repetition towards the end of longer texts. 
Finally, \texttt{diversity} offers a platform and package in which researchers from a variety of domains can use to facilitate evaluation and gather insights about diversity between human- and model-produced texts.  

\section{Limitations}
In this work, we do not attempt to measure human judgments of diversity, which are straightforward for short texts (e.g., questions) but far more difficult for longer summaries or large instruction datasets \citep{tevet-berant-2021-evaluating}; we leave this for future work. All evaluations are conducted in English.

\section*{Acknowledgements}
We gratefully acknowledge the National Science Foundation (RI 2211954) for supporting this work. We thank Ramya Namuduri for contributing the QUDSim framework to the repository, and extend our appreciation to all other contributors. 

\bibliography{custom, anthology}

\clearpage
\appendix
\section*{Appendix}
\label{sec:appendix}
\section{Run-Time Considerations}
\label{runtime}
\begin{figure}
\centering
    \includegraphics[width=0.5\textwidth]{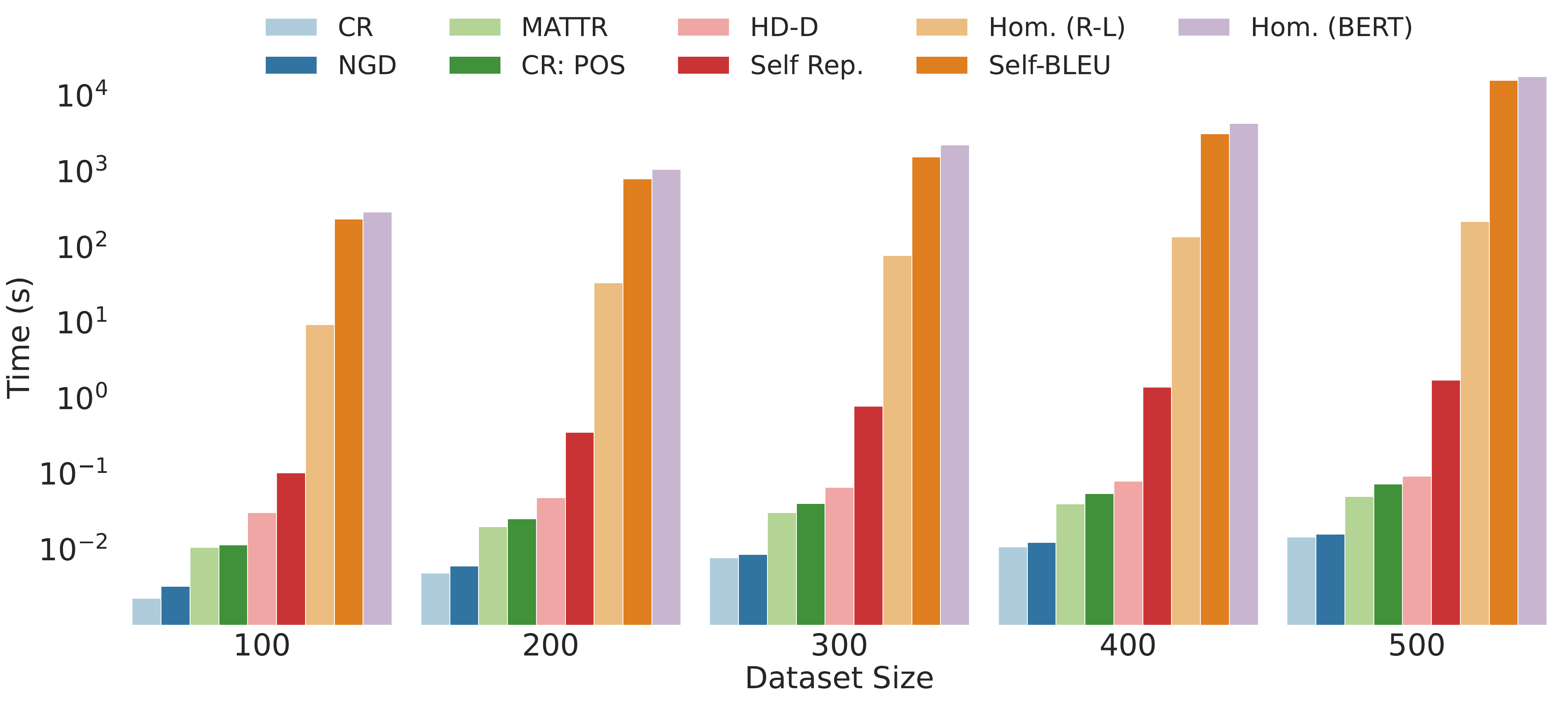}
    \caption{Mean run time \textbf{(log-scale)} on CNN/DM summaries. Run times increase with the number of text for the analysis. Even for small datasets, Self-BLEU and BERTScore homogenization are slow.} 
    \vspace{-1em}
    \label{fig:timing_ld} 
\end{figure}

Figure~\ref{fig:timing_ld} provides insights about the feasibility of obtaining scores for large samples.\footnote{Run on a single NVIDIA Quadro RTX 8000 GPU.} The compression ratio scores are fast compared to other diversity measures.

\section{Correlations with Evaluation Metrics}
\begin{figure}
\centering
    \includegraphics[width=0.35\textwidth]{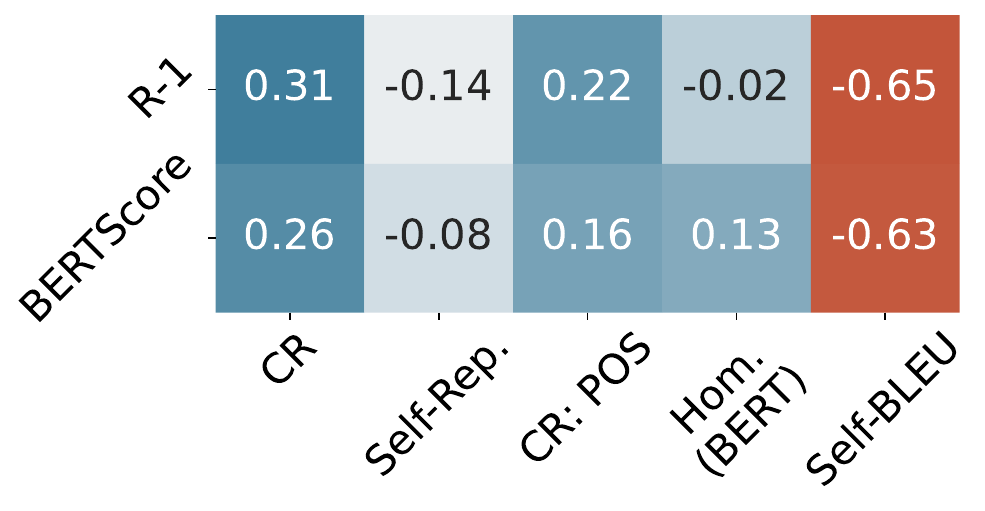}
    \caption{Correlations between diversity metrics, BERTScore, and ROUGE-1. Both reference-based metrics are weakly correlated with CR and Hom. (BERT), and moderately anti-correlated with Self-BLEU.}
    \label{fig:corr_r1_bert}
\end{figure}

Output diversity and self-repetition are aspects of model behavior that are not captured by existing evaluation approaches. 
We compute the system level correlation between the diversity scores and the traditional BERTScore and ROUGE evaluations, shown in Figure~\ref{fig:corr_r1_bert}. 

\section{Controlling for Length}
Scores on CNN/DM summaries truncated to the shortest summary reveal a different model order with respect to diversity (Table~\ref{trunc_cnn}).

\begin{table}
\begin{centering}
\resizebox{0.5\textwidth}{!}{
\begin{tabular}{llllll}
\toprule
\textbf{Model}                                                     &  \textbf{\begin{tabular}[c]{@{}l@{}}CR\\(↓) \end{tabular}}&    \textbf{\begin{tabular}[c]{@{}l@{}}CR: POS\\(↓) \end{tabular}}&\textbf{\begin{tabular}[c]{@{}l@{}}Self-\\Rep. (↓)\end{tabular}}& \textbf{\begin{tabular}[c]{@{}l@{}}Hom. \\(BERT) (↓)\end{tabular}}& \textbf{\begin{tabular}[c]{@{}l@{}}Self-\\BLEU (↓)\end{tabular}} \\\midrule
 Article                                                              & 2.268           &  5.25&2.763& 0.676                                                                &0                      \\
 Article (Lead 3) & 2.274&  5.25&2.762& 0.658&0\\
 Reference                                                             & 2.189&  5.179&2.763& 0.674                                                                         &0                      \\ \midrule
Llama-2& 2.96&    5.627&2.847& 0.674& 0.001\\
GPT-4& \textbf{2.287}&    5.376&2.761& 0.672                                                                         & \textbf{0}\\
FlanT5& 2.288&    5.389&2.779& 0.673                                                                         & \textbf{0}\\
StableLM                                                                & 2.393&    5.537&2.884& 0.672                                                                         & 0.001\\
Mistral                                                            & 2.32&    5.415&2.812& \textbf{0.67}& \textbf{0}\\
StableBeluga                                                           & 2.288&    5.46&2.766& 0.671                                                                         & \textbf{0}\\\midrule
\end{tabular}
}
\caption{Scores on CNN/DM summaries truncated 
to the shortest summary length for a given input.}
\label{trunc_cnn}
\vspace{-.75em}
\end{centering}

\end{table}

\section{Additional Evaluations}
\label{sec:broader_app} 
\paragraph{Story Writing}
\label{appendix:stories}
\begin{table}
\begin{centering}
\resizebox{0.45\textwidth}{!}{
\begin{tabular}{@{}llll}
\toprule
\textbf{Dataset} & \textbf{CR (↓)} &\textbf{CR: POS (↓)}&  \textbf{\begin{tabular}[c]{@{}l@{}}Self-Rep. (↓)\end{tabular}}\\ \midrule
 Open Assistant                          & 2.886            &6.731&3.969\\
 Unnatural Instructions                  & 4.191 &7.278&9.868\\
 Alpaca                                  & 3.119            &6.61&3.105\\
  Super-NaturalInstructions& 2.675&  5.749&3.456\\
Dolly                                   & 2.578 &6.214&  2.935\\
\midrule
\end{tabular}}
\caption{Diversity scores for instruction datasets. We do not include Self-BLEU nor Hom. (BERT) due to long run times. 
}
\label{table:instruction_diversity}
\end{centering}
\end{table}

\citet{padmakumar-etal-2023-investigating} presented an analysis of human-written stories, where people wrote either by themselves or with the help of GPT-3 or GPT-3.5 Turbo. We also find that all diversity scores agree that people writing independently produce the more diverse texts (cf. Table~\ref{appendix_table:stories_diversity}). Length is not an issue because the average length of stories in each setting are comparable: 375 words for writing without help, 372 words when writing with GPT-3 and 370 when writing with GPT-3.5.
 \begin{table*}
\begin{centering}
\resizebox{0.85\textwidth}{!}{

\begin{tabular}{@{}lllllllllll}
\toprule
\textbf{Model}& \textbf{\begin{tabular}[c]{@{}l@{}}Avg.\\Length\end{tabular}}&\textbf{\begin{tabular}[c]{@{}l@{}}CR\\(↓) \end{tabular}}& \textbf{\begin{tabular}[c]{@{}l@{}}CR: POS\\(↓) \end{tabular}}& \textbf{\begin{tabular}[c]{@{}l@{}}NGD\\(↑) \end{tabular}} &  \textbf{\begin{tabular}[c]{@{}l@{}}Self-\\Rep. (↓) \end{tabular}}&\textbf{\begin{tabular}[c]{@{}l@{}}Hom. \\(R-L) (↓)\end{tabular}}& \textbf{\begin{tabular}[c]{@{}l@{}}Hom. \\(BERT) (↓)\end{tabular}}& \textbf{\begin{tabular}[c]{@{}l@{}}Self-\\BLEU (↓)\end{tabular}} & \textbf{\begin{tabular}[c]{@{}l@{}}MATTR\\(↑) \end{tabular}} & \textbf{\begin{tabular}[c]{@{}l@{}}HD-D\\(↑) \end{tabular}} \\\midrule
 Article    &  310.20&2.511            & 5.555& 2.756            & 5.643& 0.110                                                                            & 0.695                                                                        & 0.002                  & 0.838              &0.892              \\ 
 Article (Lead-3) &  55.94&2.316            & 5.454& 3.107            & 3.999& 0.103                                                                           & 0.683                                                                        & 0                      & 0.860               &0.891              \\ 
 Reference     &  21.04                                         &2.276            & 5.409& 3.211            & 2.914& 0.081                                                                           & 0.673                                                                        & 0                      & 0.877              &0.888              \\\midrule
 StableLM  & 109.20& 2.745& 6.008& 2.636& 4.687& 0.130                                                                  & 0.695                                                                        & 0.002                  & 0.78               &0.854              \\
 Llama-2  & 102.48& 2.634            & 5.802& 2.795            & 4.618& 0.128                                                                           & 0.687                                                                        & 0.002                  & 0.795              &0.858              \\
 Mistral  & 95.18                                         & 2.531            & 5.708& 2.911            & 4.495& \underline{\emph{0.132}}& \underline{\emph{0.698}}& \underline{\emph{0.044}}                  & 0.819              &0.867              \\
 StableBeluga  & 88.46& 2.461            & 5.673& 2.992            & 4.418& 0.124                                                                           & \underline{\emph{0.698}}& \underline{\emph{0.046}}& 0.837              &0.88               \\ 
GPT-4   &  62.15&2.394            &5.531*& 3.079            &  4.041&0.104                                                                           & 0.682                                                                        & 0                      & 0.848              & 0.886              \\
FlanT5    &  20.93&\textbf{\underline{\emph{2.666}}}            &\textbf{\underline{\emph{6.222}}}& \textbf{\underline{\emph{2.743}}}            &  \textbf{\underline{\emph{2.868}}}&\textbf{\underline{\emph{0.114}}}                                                                           & 0.665                                                                        & 0.001                  & \textbf{\underline{\emph{0.756}}}& \textbf{0.842}\\ 
\midrule
\end{tabular}
}
\caption{Diversity scores for XSUM summaries. Arrow indicate the direction of more diverse texts for each score.}
\label{appendix_table:xsum_diversity}
\end{centering}

\end{table*}

\begin{table}
\begin{centering}
\resizebox{0.5\textwidth}{!}{
\begin{tabular}{@{}llllll}
\toprule
\textbf{Dataset}& \textbf{\begin{tabular}[c]{@{}l@{}}CR\\(↓) \end{tabular}}& \textbf{\begin{tabular}[c]{@{}l@{}}CR: POS\\(↓) \end{tabular}}&\textbf{\begin{tabular}[c]{@{}l@{}}Self-\\Rep. (↓)\end{tabular}}& \textbf{\begin{tabular}[c]{@{}l@{}}Hom. \\(BERT) (↓)\end{tabular}}&  \textbf{\begin{tabular}[c]{@{}l@{}}Self-\\BLEU (↓)\end{tabular}} \\ \midrule
Solo& 2.901& 5.314&5.873& 0.604& 0.018\\
 GPT-3& 2.940&  5.371&5.911& 0.613&0.020               \\
InstructGPT& 3.064& 5.462&5.966& 0.631& 0.022\\\midrule
\end{tabular}
}

\caption{Diversity scores over essays. Working with an LLM correlates with lower diversity.}
\label{appendix_table:stories_diversity}
\end{centering}
\end{table} 

\begin{figure}
\centering
    \includegraphics[width=0.4\textwidth]{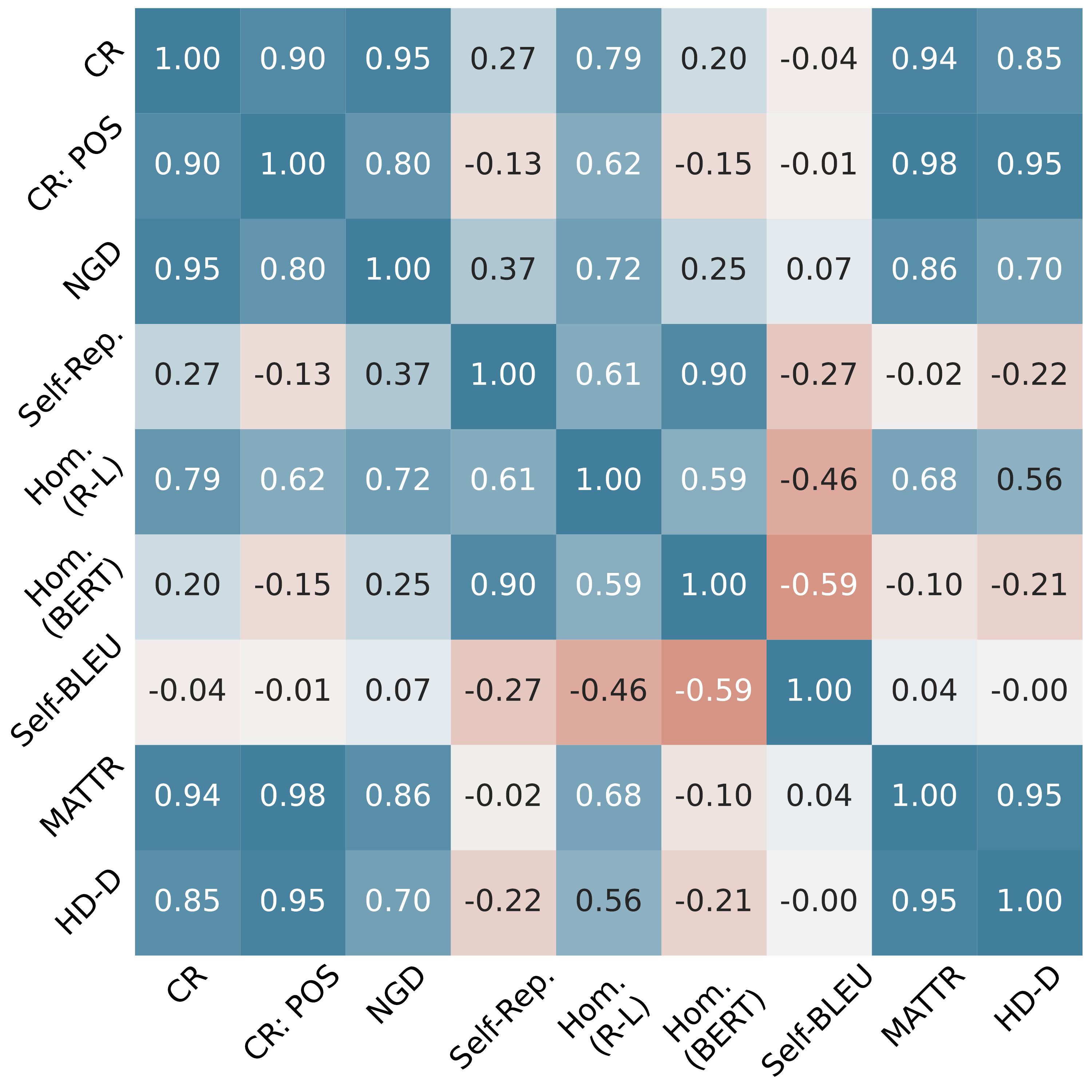}
    \caption{Correlation table between scores on XSUM.}
    \label{fig:corr_xsum}
\end{figure}

\paragraph{Instruction-tuning Datasets}
\label{appendix:instructions}

The quality and diversity of instructions are likely to result in more robust and capable systems \citep{DBLP:conf/iclr/SanhWRBSACSRDBX22,mishra-etal-2022-cross}. We analyze the diversity of five instruction-tuning datasets: Open Assistant \citep{kopf2024openassistant}, Super-NaturalInstructions \citep{wang-etal-2022-super}, Unnatural Instructions \citep{honovich-etal-2023-unnatural}, Alpaca \citep{wang-etal-2023-self-instruct}, and Dolly \citep{DatabricksBlog2023DollyV2} (Table~\ref{table:instruction_diversity}). 

Open Assistant instructions are remarkably diverse compared to the other datasets across all diversity scores. Unnatural instructions are remarkable in the opposite direction. We provide an analysis of the diversity scores with the length controlled in Appendix~\ref{appendix:instructions_length}.
\begin{table}
\begin{centering}
\resizebox{0.45\textwidth}{!}{
\begin{tabular}{llllll}
\midrule
\textbf{Model}& \multicolumn{1}{l}{\textbf{\begin{tabular}[c]{@{}l@{}}CR\\(↓) \end{tabular}}} &  \textbf{\begin{tabular}[c]{@{}l@{}}CR: POS\\(↓) \end{tabular}}&\multicolumn{1}{l}{\textbf{\begin{tabular}[c]{@{}l@{}}Self-\\Rep. (↓)\end{tabular}}}  & \multicolumn{1}{l}{\textbf{\begin{tabular}[c]{@{}l@{}}Hom. \\(BERT) (↓)\end{tabular}}} & \multicolumn{1}{l}{\textbf{\begin{tabular}[c]{@{}l@{}}Self-\\BLEU (↓)\end{tabular}} }   \\\midrule
 Article                                 & 2.162                                                       &  5.095&2.719& 0.666                                                                                                                    &0                                                                  \\
 Article (Lead 3)& 2.179&  5.093&2.719& 0.663&0\\
 Reference                               & 2.230                                                        &  5.314&2.663& 0.667                                                                                                                    &0                                                                  \\ \midrule
Llama-2                                 & 2.345                                                       &  5.636&2.919& 0.663                                                                                                                    & 0.002\\
GPT-4                                   & 2.213                                                       &  5.425&2.666& 0.663                                                                                                                    & 0                                                                  \\
FlanT5                                  & 2.490&  5.737&2.707& 0.665& 0.001                                                              \\
StableLM                                & 2.342                                                       &  5.521&2.823& 0.664                                                                                                                    & 0.001                                                              \\
Mistral                                 & 2.308                                                       &  5.689&2.736& 0.659                                                                                                                    & 0                                                                  \\
StableBeluga                            & 2.210                                                        &  5.436&2.663& 0.659                                                                                                                    & 0                                                                  \\\midrule
\end{tabular}
}
\caption{Diversity metrics for XSUM summaries, with outputs from each model truncated to the length of the shortest. All scores are directly comparable.}
\label{appendix_table:xsum_diversity_length_controlled}
\end{centering}

\end{table}

 Given the large dataset sizes, ranging from 15-80k data points, we do not compute the homogenization scores nor Self-BLEU, as the computation time is infeasible. For approximately 50k instructions, the estimated computation times ranged from 48 to 800 hours for these scores. This case study highlights the relevancy of the run-time analysis for computing score that we presented in the previous section.



\subsection{Correlation Between Metrics}
Self-BLEU scores are almost perfectly correlated between the two datasets; they appear to not be affected by text source. The other scores are still moderately to highly correlated but as already observed, models are ranked differently. When reporting diversity, source of analyzed data also has to be taken into account, in addition to length. 


\subsection{Instruction Datasets, Length Controlled}
\label{appendix:instructions_length}
 Table \ref{table:instruction_diversity_length_controlled} shows scores for instructions downsampled to the size of the smallest dataset, and truncated to the length of the shortest instructions in the remaining data. Again, the Open Assistant dataset stand out as most diverse, while the Unnatural Instructions dataset is markedly less diverse than the others. Self-repetition in the related Super-Natural and Unnatural instructions is notably high. The human instructions in Dolly compare favorably with automatic instructions, especially when bearing in mind that only eight tasks are covered in it. CR:POS points to Super-natural instructions as the most diverse. We do not have a convincing explanation of why it compares so favorably against others on this score.
\begin{table}
\begin{centering}
\resizebox{0.5\textwidth}{!}{
\begin{tabular}{@{}llll}
\toprule
\textbf{Dataset} & \textbf{CR (↓)} &\textbf{CR: POS (↓)}&  \textbf{\begin{tabular}[c]{@{}l@{}}Self-Rep. (↓)\end{tabular}}\\ \midrule
 Open Assistant                          & 2.370&5.402&1.741\\
 Unnatural Instructions                  & 6.036&8.421&5.595\\
 Alpaca                                  & 3.301&6.044&2.020\\
 Super-NaturalInstructions& 2.458&  1.844&4.859\\
Dolly                                   & 2.832&5.504&  2.235\\
\midrule
\end{tabular}
}
\caption{Truncated diversity scores for instruction datasets. }
\label{table:instruction_diversity_length_controlled}
\end{centering}
\end{table}

\subsection{XSUM Metrics}

Tables~\ref{appendix_table:xsum_diversity},~\ref{appendix_table:xsum_diversity_length_controlled} show the full diversity metrics over XSUM with and without controlling for length. 

Figure~\ref{fig:corr_xsum} shows the correlations between all pairs of metrics for the XSUM dataset. The correlations show that  compression ratio is highly to moderately correlated with other n-gram scores, similar to the findings for the CNN/DM dataset.

\end{document}